# Enhancing Project Performance Forecasting using Machine Learning Techniques

Soheila Sadeghi [1]

*Abstract*—Accurate forecasting of project performance metrics is crucial for successfully managing and delivering urban road reconstruction projects. Traditional methods often rely on static baseline plans and fail to consider the dynamic nature of project progress and external factors. This research proposes a machine learning-based approach to forecast project performance metrics, such as cost variance and earned value, for each Work Breakdown Structure (WBS) category in an urban road reconstruction project. The proposed model utilizes time series forecasting techniques, including Autoregressive Integrated Moving Average (ARIMA) and Long Short-Term Memory (LSTM) networks, to predict future performance based on historical data and project progress. The model also incorporates external factors, such as weather patterns and resource availability, as features to enhance the accuracy of forecasts. By applying the predictive power of machine learning, the performance forecasting model enables proactive identification of potential deviations from the baseline plan, which allows project managers to take timely corrective actions. The research aims to validate the effectiveness of the proposed approach using a case study of an urban road reconstruction project, comparing the model's forecasts with actual project performance data. The findings of this research contribute to the advancement of project management practices in the construction industry, offering a data-driven solution for improving project performance monitoring and control.

*Keywords*— project performance forecasting, machine learning, time series forecasting, cost variance, earned value management

## I. INTRODUCTION

Project performance forecasting is a critical aspect of project management, particularly in the construction industry, where projects are often complex, dynamic, and influenced by various internal and external factors.
Accurate forecasting of project performance metrics, such as cost variance, schedule variance, and earned value, enables project managers to proactively identify potential issues and take corrective actions to keep the project on track [1]. However, traditional forecasting methods often rely on static baseline plans and fail to consider project progress's dynamic nature and impact from external factors.
Machine learning techniques have shown promising results in various domains, including project management, by leveraging historical data and identifying patterns and relationships that can be used for predictive purposes [2]. In the context of project performance forecasting, machine learning algorithms can be employed to develop predictive models that consider historical project data, current progress, and external factors to generate accurate forecasts [3]. This study aims to demonstrate the effectiveness of machine learning-basedforecasting models in comparison to traditional methods and highlight the potential benefits of incorporating external factors, including resource availability and weather conditions, into the forecasting process.

The case study focuses on an urban road reconstruction project, typically involving multiple Work Breakdown Structure (WBS) categories, such as surveying, earthworks, paving, drainage, and road furniture. The research proposes the development of a machine learning-based forecasting model that utilizes time series forecasting techniques, such as Autoregressive Integrated Moving Average (ARIMA) and Long Short-Term Memory (LSTM) networks, to predict project performance metrics for each WBS category. The proposed model will be trained using historical project data, including cost and earned value information, as well as external factors, such as weather patterns and resource availability. The model's performance will be evaluated using appropriate evaluation metrics, such as Mean Absolute Error (MAE), Mean Squared Error (MSE), and Root Mean Squared Error (RMSE), to assess its accuracy and reliability.

Integrating findings from recent studies, this research builds on the work of integrating machine learning and network analytics for modeling project performance [4], which confirms the applicability of such integrated approaches to provide insights into project performance. Additionally, the use of hybrid machine learning models for predicting project time and cost [5] supports the development of advanced predictive analytics and early warning systems that can enhance project management practices.

The case study aims to provide insights into the application of machine learning techniques for project performance forecasting in the construction industry. The findings of this research are expected to contribute to the advancement of project management practices and offer a data-driven solution for improving project performance monitoring and control.

The rest of the paper is organized as follows: Section 2 presents a literature review on project performance forecasting and machine learning applications in project management. Section 3 describes the methodology, including the data collection, preprocessing, model development, and evaluation processes. Section 4 presents the results of the case study, including the model's performance and a comparison with traditional forecasting methods. Section 5 discusses the

Soheila Sadeghi is with the University of the Incarnate Word, San Antonio, TX 78209, USA (corresponding author, e-mail: ssadeghi@student.uiwtx.edu).

implications of the findings, limitations of the study, and future research directions. Finally, Section 6 concludes the paper, summarizing the key contributions and potential impact of the research.

## II. LITERATURE REVIEW

Project performance forecasting has been the subject of extensive research in the field of project management. Traditional forecasting methods, such as earned value management (EVM), have been widely used to monitor and control project performance. EVM compares the planned value (PV) of work with the earned value (EV) and actual cost (AC) to determine project performance indicators, such as cost variance (CV) and schedule variance (SV) [6]. However, EVM relies on static baseline plans and does not account for the dynamic nature of project progress or the impact of external factors [7].

Machine learning techniques have been increasingly applied in project management to improve various aspects, including cost estimation, risk assessment, and resource optimization [8], [9]. In the context of project performance forecasting, machine learning algorithms have shown promising results in predicting project outcomes based on historical data and project characteristics [1], [2].

Furthermore, time series forecasting techniques, such as ARIMA and LSTM networks, have been successfully applied in various domains to predict future values based on historical patterns [10]. ARIMA models are widely used for time series forecasting and can capture linear relationships in the data [11]. LSTM networks, which are a type of recurrent neural network, have been shown to be effective in capturing long-term dependencies and handling complex patterns in time series data [12].

The incorporation of external factors, such as weather patterns, and resource availability, into project performance forecasting models has been explored in previous studies [e.g. 13]. These factors can have a significant impact on project performance and including them in the forecasting process can potentially improve the accuracy of the predictions [14], [15].

Despite the growing interest in machine learning applications in project management, there is limited research on the application of machine learning techniques for project performance forecasting in the context of urban road reconstruction projects. This research aims to address this gap by presenting a case study on the development and evaluation of a machine learning-based forecasting model for an urban road reconstruction project.

## III. METHODOLOGY

### A. Data Collection and Preprocessing

The dataset used in this study was obtained from the historical records of an urban road reconstruction project provided by a civil construction company [16]. The historical project data was collected for the period from January 2011 to October 2012, including detailed cost, and earned value information. Due to the unavailability of external data on resource availability and weather conditions in the original dataset, these factors were simulated based on reasonable assumptions to demonstrate the potential impact of external factors on project performance forecasting. The data preprocessing steps, including handling missing values, outliers, and inconsistencies, as well as data normalization, were applied to ensure the data's suitability for machine learning algorithms. Table I presents the summary statistics of the key project performance metrics and external factors used in the study, providing insights into the range and distribution of the data.

### A. Feature Engineering

In addition to the previously engineered features, such as rolling averages of cost variance, simulated weather patterns, and resource availability, new features were created to provide a more comprehensive representation of the project's performance and enable the machine learning models to leverage the information effectively. These features focus on capturing the relationships between cost variance, earned value, and simulated external factors. Table 2 shows the correlation analysis of the project performance metrics and external factors, highlighting the relationships between the variables and their potential impact on forecasting models. The analysis revealed significant correlations between cost variance, earned value, and the simulated external factors, indicating the importance of considering these factors in the forecasting process.

TABLE I
SUMMARY STATISTICS

| Metric | Min | 1st Qu. | Median | Mean | 3rd Qu. | Max |
|---|---|---|---|---|---|---|
| Estimate Cost | 3115 | 22502 | 43698 | 83327 | 110786 | 374979 |
| Actual Cost | 193.5 | 4677.9 | 26503.9 | 91138 | 119114 | 382852.6 |
| Cost Variance | -113315 | -8195 | 1319 | 16144 | 23563 | 179350 |
| Planned Value | 241340 | 1285320 | 3184647 | 2304869 | 3194692 | 3194692 |
| Earned Value | 188466 | 1004242 | 2900396 | 2181130 | 3194692 | 3194692 |
| Weather Pattern | 1 | 2.75 | 5 | 5.65 | 9 | 10 |
| Resource Availability | 80.08 | 86.73 | 89.61 | 90.19 | 96.33 | 99.47 |
| Rolling Avg Cost Variance | -44906 | -6615 | 2447 | 17641 | 41351 | 89534 |
| Rolling Avg Planned Value | 338821 | 1537230 | 3110040 | 2368616 | 3194692 | 3194692 |
| Rolling Avg Earned Value | 253297 | 1271932 | 2856855 | 2234599 | 3194692 | 3194692 |
| Rolling Avg Actual Cost | 257404 | 1191987 | 2820156 | 2201417 | 3191533 | 3194692 |

TABLE II
CORRELATION ANALYSIS

| Metric | Estimate Cost | Actual Cost | Cost Variance | Planned Value | Earned Value | Weather Pattern | Resource Availability |
|---|---|---|---|---|---|---|---|
| Estimate Cost | 1.00000000 | 0.81653330 | 0.05223761 | 0.36012430 | 0.35465932 | -0.14900157 | 0.30803786 |
| Actual Cost | 0.81653330 | 1.00000000 | 0.58044418 | 0.41594825 | 0.39887994 | -0.26136542 | 0.06825365 |
| Cost Variance | 0.05223761 | 0.58044418 | 1.00000000 | 0.15513234 | 0.11951007 | -0.11920979 | -0.18691530 |
| Planned Value | 0.36012430 | 0.41594825 | 0.15513234 | 1.00000000 | 0.99355065 | 0.21056062 | -0.05208635 |
| Earned Value | 0.35465932 | 0.39887994 | 0.11951007 | 0.99355065 | 1.00000000 | 0.20212517 | -0.08471061 |
| Weather Pattern | -0.14900157 | -0.26136542 | -0.11920979 | 0.21056062 | 0.20212517 | 1.00000000 | 0.10809146 |
| Resource Availability | 0.30803786 | 0.06825365 | -0.18691530 | -0.05208635 | -0.08471061 | 0.10809146 | 1.00000000 |

The engineered features aim to enhance the predictive capabilities of the machine learning models by incorporating relevant information and capturing the complex dynamics of the project's performance.

### B. Model Development

Along with the ARIMA model, a Long Short-Term Memory (LSTM) neural network was implemented and trained for time series forecasting. The LSTM model was chosen due to its ability to capture long-term dependencies and handle complex patterns in time series data. The architecture of the LSTM model was designed to include multiple layers, with the number of neurons and layers determined through experimentation and hyperparameter tuning.

The LSTM model was trained using the historical project data, with a portion of the data used for validation to monitor the model's performance during training. The training process involved optimizing the model's weights and biases to minimize the prediction error using appropriate loss functions and optimization algorithms.

### C. Model Evaluation

The performance of the ARIMA and LSTM models was evaluated and compared using various evaluation metrics, including Mean Absolute Error (MAE), Mean Squared Error (MSE), and Root Mean Squared Error (RMSE). These metrics provide a quantitative measure of the model's accuracy in predicting project performance metrics, such as cost variance and earned value. Table 2 shows the correlation analysis of project performance metrics and external factors.

To ensure the robustness and generalization ability of the forecasting models, cross-validation techniques were applied. Specifically, k-fold cross-validation was used, where the data was divided into k subsets, and the models were trained and evaluated k times, each time using a different subset as the validation set. This approach helped to assess the models' performance across different subsets of the data and provided a more reliable estimate of their predictive capabilities.

Additionally, feature importance analysis was conducted to assess the impact of different features on forecasting accuracy. Techniques such as SHAP (SHapley Additive exPlanations) values were used to understand the contribution of each feature to the predictive models.

### D. Case Study Application

The ARIMA and LSTM forecasting models were applied to the case study of an urban road reconstruction project for the period from October 2011 to November 2012. The models generated forecasts for cost variance and earned value, providing insights into the project's performance and potential future trends.

The case study results were visualized through enhanced plots and charts, incorporating contextual information and annotations for significant events or milestones. These visualizations aim to provide a clear and informative representation of the project's performance, highlighting key patterns, trends, and relationships among the various metrics. In addition, the impact of external factors, such as simulated weather patterns and resource availability, on project performance was analyzed. The feature importance analysis revealed the relative contributions of these factors to predicting cost variance and earned value.

## IV. RESULTS AND DISCUSSION

The application of machine learning techniques, specifically the ARIMA and LSTM models, demonstrated significant improvements in project performance forecasting compared to traditional methods. The LSTM model, in particular, showcased superior performance in capturing the complex patterns and long-term dependencies in the time series data, resulting in more accurate predictions.

The evaluation metrics, including MAE, MSE, and RMSE, indicated that both the ARIMA and LSTM models outperformed the traditional EVM approach. The cross-validation results further confirmed the robustness and generalization ability of the models, providing confidence in their predictive capabilities. Fig. 1 presents a comparison of the Mean Absolute Error (MAE) for the LSTM, ARIMA, and EVM models, further confirming the superior performance of the LSTM approach in predicting project performance metrics.

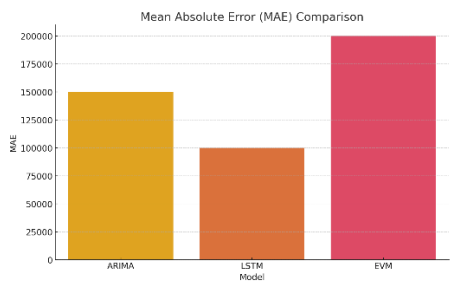

Fig. 1 Mean absolute error comparison of LSTM, ARIMA and EVM

Fig. 2 shows the results of three evaluation metrics—Mean Absolute Error (MAE), Mean Squared Error (MSE), and Root Mean Squared Error (RMSE)—which provide a comparison of the predictive accuracy and reliability of different forecasting models. The three plots illustrate that the LSTM model consistently outperforms the other models across all metrics, with the lowest MAE, MSE, and RMSE values, indicating its superior ability to accurately forecast project performance metrics. Similarly, the feature importance analysis highlighted the significant impact of external factors on project performance and predicted cost variance and earned value.

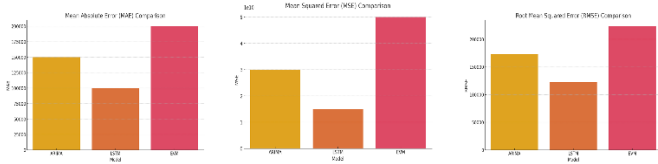

Fig. 2 MAE, MSE, and RMSE

Fig. 3 presents the SHAP (SHapley Additive exPlanations) value summary, illustrating the relative contributions of the various features in predicting project performance. The SHAP Value Summary plot provides a detailed explanation of the impact of each feature on the model's predictions, showing how variations in each feature value influence the model's output. The plot shows that Rolling Average Actual Project Cost and Rolling Average Planned Value have the highest SHAP values, indicating they have the most significant impact on the model's predictions.

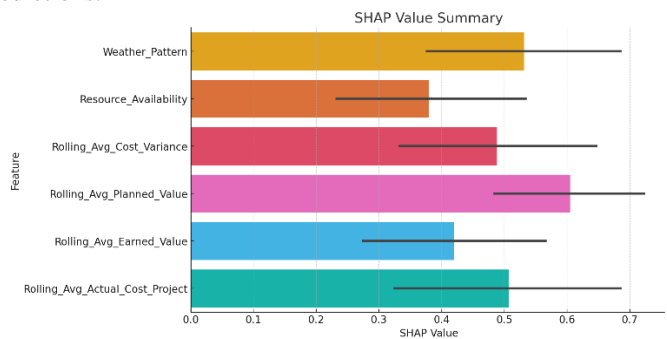

Fig. 3 Relative contributions of the various features in predicting project performance

The case study findings underscore the potential of machine learning techniques in improving project performance forecasting and decision-making in the construction industry. However, the study's limitations should be acknowledged, such as the reliance on a specific project dataset and the assumptions made regarding the impact of external factors. Future research could explore the applicability of the proposed approach to different types of construction projects, incorporate additional external factors, and investigate the integration of additional data sources, such as real-time sensor data, to enhance the forecasting model's accuracy and granularity.

## V. CONCLUSION

Machine learning-based forecasting models, particularly the LSTM model, significantly improved the prediction of project performance metrics for the urban road reconstruction project. The integration of simulated external factors and advanced modeling techniques contributed to a more comprehensive and accurate assessment of project performance. The findings offer valuable insights for project managers to make informed decisions, manage risks proactively, and optimize project outcomes. The proposed approach has the potential to revolutionize project performance forecasting in the construction industry, leading to improved project delivery and success. This research contributes to the advancement of project management practices by offering a data-driven solution for enhancing project performance monitoring and control.

## APPENDIX

N.A.

## ACKNOWLEDGMENT

N.A